# Robust feature knowledge distillation for enhanced performance of lightweight crack segmentation models

Zhaohui Chen [a], Elyas Asadi Shamsabadi [a], Sheng Jiang [b], Luming Shen [a], Daniel Dias-da-Costa [a,*]

[a] *School of Civil Engineering, Faculty of Engineering, The University of Sydney, Sydney, NSW 2006, Australia*

[b] *College of Water Conservancy and Hydropower Engineering, Hohai University, Nanjing, 210098, China*

**Abstract:**

Vision-based crack detection faces deployment challenges due to the size of robust models and edge device limitations. These can be addressed with lightweight models trained with knowledge distillation (KD). However, state-of-the-art (SOTA) KD methods compromise anti-noise robustness. This paper develops Robust Feature Knowledge Distillation (RFKD), a framework to improve robustness while retaining the precision of light models for crack segmentation. RFKD distils knowledge from a teacher model's logit layers and intermediate feature maps while leveraging mixed clean and noisy images to transfer robust patterns to the student model, improving its precision, generalisation, and anti-noise performance. To validate the proposed RFKD, a lightweight crack segmentation model, PoolingCrack Tiny (PCT), with only 0.5 M parameters, is also designed and used as the student to run the framework. The results show a significant enhancement in noisy images, with RFKD reaching a 62% enhanced mean Dice score (mDS) compared to SOTA KD methods.

**Keywords:** Crack Segmentation; Knowledge Distillation; Transformer; Lightweight Network, Noise Robustness

**Paper Highlights:**

- A novel knowledge distillation method is developed for robust crack segmentation.
- A novel Transformer-based student model with only 0.5 M parameters is designed.
- Knowledge is distilled from logit layers and intermediate feature maps.
- Robust knowledge is imparted through training mixed clean and noisy data.
- The proposed method exhibits up to 62 % improved mDS against noisy data.

## 1. Introduction

Crack detection is crucial in preserving the structural integrity of various infrastructures. The appearance of cracks may indicate aging, deterioration, or internal structural issues, which may cause major failures if not detected early [1]. Structural elements are prone to cracking under various combinations of loads, resulting in potential safety risks and reduced structural lifespan [2]. Despite the importance of continuous monitoring and inspection of structures [3,4], traditional manual methods are discussed to have various drawbacks, mainly in the aspects of efficiency and safety [5-7]. Recent advanced technologies have enabled automatic crack detection [8,9]; in particular, the emergence of deep learning [10] has remarkably enhanced autonomous crack detection [11].

Convolutional neural networks (CNNs) [12-16] were pioneers in classical deep learning for computer vision-based automatic crack detection. Due to the limitation caused by a focus of CNNs on local dependencies, Transformers, with the ability to capture multi-range information in data, were proposed to overcome this drawback and improve image recognition accuracy [17-20]. Following the introduction of Transformers in crack segmentation, significant accuracy and generalisation have been achieved, yet practical deployment still faces challenges due to the large scale of Transformer-based models and the restricted computational power of edge devices [21]. As lightweight deep-learning models were continuously proposed to address the issue of deployment on edge devices, crack recognition precision of current light models and limited resolution image input due to current restricted computational power still hindered their further application [22]. For higher image resolution input on edge devices, more compact models are yet to be developed [23]. As for the performance improvement of lightweight models, one effective solution is distilling knowledge from the heavyweight to smaller models to improve the patterns learnt during the training process [24].

In a KD process, the student model learns from both the ground truth (GT) and the soft targets provided by the teacher model [24]. By mimicking the behaviour of the teacher, the student model distils the knowledge encapsulated in the teacher's predictions, thereby achieving comparable performance [25]. Unlike traditional compression methods such as model pruning [26] or low-rank factorisation [27], KD allows both the smaller student model and the larger teacher model to have similar or different network structures [28,29]. Thus, the teacher and student models can be built in various ways, providing versatility in how models are designed. Whether the models share identical architectures or diverge significantly, KD accommodates the transfer of knowledge seamlessly, allowing for tailored solutions across diverse tasks and domain [28]. This adaptability is a defining feature of KD, enabling researchers and practitioners to explore a wide range of network configurations and design choices without sacrificing performance [30].

While numerous studies in the literature are dedicated to the development of models for supervised crack segmentation, literature regarding KD-based crack segmentation remains sparse. Chen et al. [31] proposed a CNN-based lightweight model with only 1.18 M parameters for crack segmentation and improved the model performance through hyperthermal and non-isothermal KD strategies. This lightweight model was then demonstrated suitable for an edge device. Wang et al. [32] designed a CNN-based light model with 5.92 M parameters and enhanced model accuracy through channel-wise KD strategies [33]. The applications of KD have achieved comparable results for crack segmentation [34], however, conveying knowledge only from the logit layer can only transfer accuracy without imparting anti-noise robustness to the student models. Moreover, since current CNN-based lightweight student models are weaker against image noises than Transformer-based models [17,18], this intrinsic shortage may further restrict the practical application of these lightweight models.

Considering the limitation of existing KD strategies and CNN-based student models, this work aims to achieve robustness in knowledge transfer and design a Transformer-based lightweight model with accuracy and robustness for possible practical applications. By distilling additional knowledge from intermediate feature layers with mixed clean and noise image input, we propose a novel framework, named robust feature knowledge distillation (RFKD), for the distillation of robust patterns from a teacher to a student. Considering the need for a powerful and precise lightweight Transformer-based model which is missing in the literature, a novel Transformer-based lightweight crack segmentation model named PoolingCrackTiny (PCT) is also developed as part of this work. The core hypothesis is that the student model can learn robust crack features under the supervision of the teacher model when exposed to both clean and noisy data. Additional knowledge from the feature layers of the teacher model can help the student learn precise crack feature representations, resulting in the improvement of generalisation performance for the practical deployment.

This paper is organised as follows. Section 2 introduces the specific architectures of the proposed RFKD framework and PCT lightweight model. Section 3 illustrates the experiment implementation, including the dataset information, training settings, and assessment flowchart. In Section 4, the comprehensive results of the experiments are showcased and thoroughly discussed. Finally, the conclusions are drawn in Section 5.

## 2. Methodology

In this section, a novel robust knowledge distillation framework for crack segmentation was developed to improve both the accuracy and robustness of the distilled student model, followed by a designed student model for better leveraging the advance of the proposed framework and verifying its practical application.

### 2.1 RFKD framework

Currently, KD strategies for crack segmentation are based on transferring knowledge between the logit layers of teacher and student models [31,32]. In the training process, the lightweight student model needs to minimise the difference between model output and two types of labels: soft and hard [31]. The soft label is the output of the heavyweight teacher model corresponding to the current image input, and the hard label is the ground truth of the current image input. Yet, these logit layer-based KD methods have limited influence on the performance of the student model. Since feature representation learning was introduced by Bengio et al. [35], feature-based knowledge from the intermediate layers has become a reasonable extension of logit layer-based KD to supervise the training of the student model [36,37]. Yet, these KD methods can only improve the accuracy of the student model without inheriting the anti-noise robustness of the teacher model. Inspired by the adversarial training with specific noises to enhance model robustness against these designated noises [38,39], we designed a novel framework (see Fig. 1) that combines clean and noisy images as the input and distil additional intermediate feature representations of the teacher for the student model to learn robust patterns.

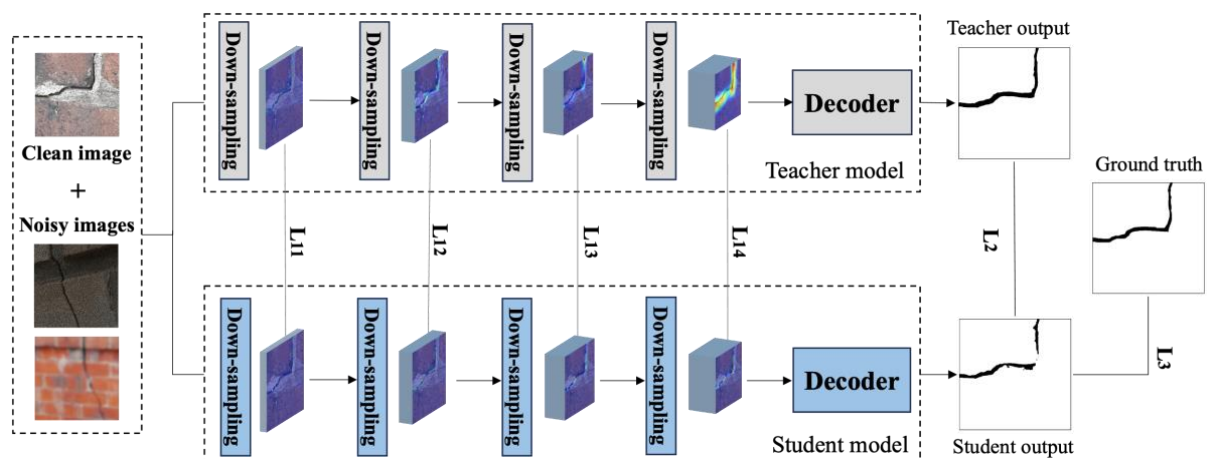

**Fig. 1** Framework of the proposed robust feature knowledge distillation (RFKD).

In this study, four common image noises, including pepper, salt, crack texture and gaussian blur, are utilised to distort parts of the clean images in the dataset. The ratio of clean and noisy images is finally established as 2:1, with an equal quantity of different types of noisy images.

The ground truth labels are first used to train the teacher model. In the next stage, as shown in Fig. 1, the student model training is under the supervision of three parts: intermediate feature maps of the trained teacher model, the final output of the trained teacher model, and the ground truth labels. $L_{1i}$ ($i$ =1, 2, 3, 4), $L_2$, and $L_3$ are the respective losses for these three parts, and the total training loss is formulated as:

$$L = \sum \alpha_i \cdot L_{1i} + \beta \cdot L_2 + \gamma \cdot L_3, \quad (1)$$

where $\alpha_i$ ($i$ =1, 2, 3, 4), $\beta$ and $\gamma$ are hyperparameters, representing the weight of each part, that are set before training. In the first two items of the total loss, to maximise the similarity of the student and teacher model outputs, the pair-wise distance loss [37] was utilised to minimise the pixel-level Euclidean distance of their intermediate and final outputs. Obtaining feature map tensors from the intermediate and logits layer of the student and teacher models, the pair-wise distance loss $L^{pw}$ is formulated as follows:

$$L^{pw} = \|F^S - F^T\|, \quad (2)$$

here $F^S$ and $F^T$ denote the feature map tensors of the student and the teacher models, and $\|\cdot\|$ is the Frobenius norm [40].

For the student model training with the ground truth labels in the last item of the total loss, Dice loss [41] was chosen due to its balanced performance in image segmentation. This loss function ($L^{DS}$) is derived from the Dice score (DS), as shown in the following:

$$L^{DS} = 1 - DS, \quad (3)$$

$$DS = \frac{2|P \cap T|}{|P| + |T|}, \quad (4)$$

where $P$ represents the predicted pixel map, and $T$ denotes the labelled pixel map. According to Eq. (1), the total loss for the student model training is:

$$L = \sum \alpha_i \cdot L_{1i}^{pw} + \beta \cdot L_2^{pw} + \gamma \cdot L_3^{DS}, \quad (5)$$

where the pair-wise distance loss is applied for the soft label distillation from the teacher ($L_{1i}^{pw}$ and $L_2^{pw}$), and Dice loss is selected for the hard label distillation from the ground truth ($L_3^{DS}$). To consider each loss item equally, the hyperparameters are set as: $\alpha_i = 0.25$ ($i$ =1, 2, 3, 4), $\beta = 1$, $\gamma = 1$.

Compared with previous KD strategies [31,32] only considering $L_2$ and $L_3$, RFKD forces the student model to mimic each feature map of the teacher model after down-sampling, leading to the gradually grasping feature representation of the teacher model in the inference progress. In addition, to address the issue of non-robust KD, clean and noisy images are mixed as the input for the student model training. This mixed image input acts as straightforward adversarial training for the student model, assisting the student in imitating the robust teacher when encountering image noises.

## 2.2 PCT architecture

To demonstrate the advantage of the RFKD-based trained student model in practical deployment, we designed a Transformer-based student model with only 0.5 M parameters. The overall structure of the proposed lightweight PoolingCrackTiny is illustrated in Fig. 2, built upon the PoolingCrack model [42] (See Appendix-A.2).

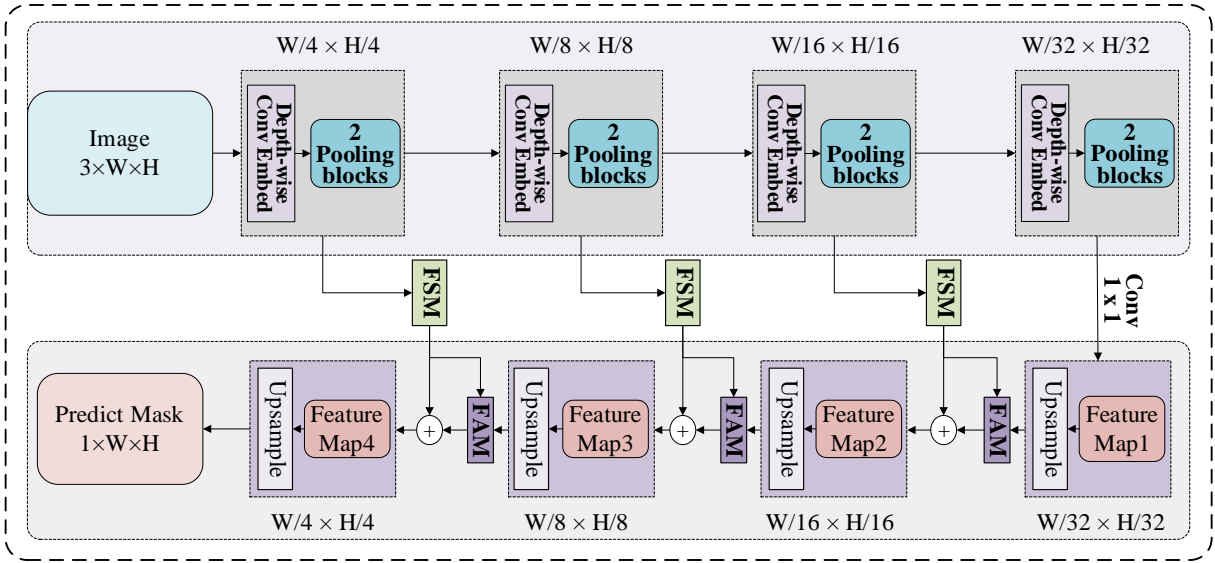

**Fig. 2** Architecture of the proposed lightweight model PoolingCrackTiny (PCT).

Compared with PoolingCrack, the number of pooling blocks and patch embedding channels were respectively reduced to [2, 2, 2, 2] and [48, 96, 192, 384] in the PCT model. Additionally, depth-wise convolution [43] was applied in the convolutional patch embedding (See Fig. 2) due to its effectiveness in shrinking the model scale. The feature selection module (FSM) [44] and feature alignment module (FAM) [45] in PoolingCrack were kept in PCT to maintain its overall performance.

## 3. Study design

In this section, the representative KD strategies and lightweight models were selected to run the comparison, thus validating the performance of the proposed RFKD framework and PCT model. The evaluation metrics, image dataset, training configuration, and study framework are also discussed in detail.

## 3.1 KD strategy and lightweight model selection

To verify the performance of the proposed RFKD method and evaluate the capabilities of the distilled PCT lightweight model, two comparisons were conducted respectively with the representative KD strategies and lightweight models. Naïve Knowledge Distillation (NKD) [24], Channel-wise Knowledge Distillation (CWD) [33], and DIST [46] were chosen for the KD strategy comparison. CNN-based MobileNetV2 [47], MobileOne [48], EfficientNet-B0 [49], and Transformer-based MobileViT [50] were also selected as established and SOTA CNN- and Transformer-based models to analyse the performance of the lightweight PCT model, designed in this work.

## 3.2 Evaluation metrics

Since a crack segmentation model aims to distinguish the crack regions and the background in the crack images, two common evaluation metrics are frequently utilised to appraise the performance of the student model with KD strategies, including mean Dice score and mean intersection over union (mIoU) score. DS metric is introduced in Eq. (4), and the IoU metric can be formulated as:

$$IoU = \frac{|P \cap T|}{|P \cup T|}, \qquad (6)$$

similar to DS metric, P is the predicted pixel map, while T is the labelled pixel map. As these two metrics are insensitive to the specific threshold used to crack pixel classification and evaluate the overlap of the entire region, including the boundary, they provide a more balanced evaluation of segmentation performance than Precision and Recall, alleviating the issues of threshold dependence and boundary bias [51,52].

## 3.3 Image datasets

To assess the student model performance with different KD strategies, a crack image dataset MC448 [18] was selected for the validation experiment. This dataset was selected primarily due to the complexity of the masonry surface, which poses a greater challenge for computer vision-based frameworks compared to asphalt and concrete surfaces. The dataset poses significant challenges for crack segmentation due to the inclusion of non-crack images and the presence of crack-like textures in the background of masonry surfaces. The dataset comprises 3351 masonry crack images, each with a resolution of 448 × 448 pixels, along with their respective masks. These images were sourced from various origins, including 45 copyright-free images, 153 high-resolution manually captured photographs, and 15 artificially generated images. Since four image noises were chosen in this study for evaluating the proposed RFKD framework, 3000 images and their corresponding masks including 2000 clean and 1000 noisy images (250 images for each type of noise) were allocated for training, reserving the remaining 351 images for model evaluation. Some image samples with their masks are shown in Fig. 3.

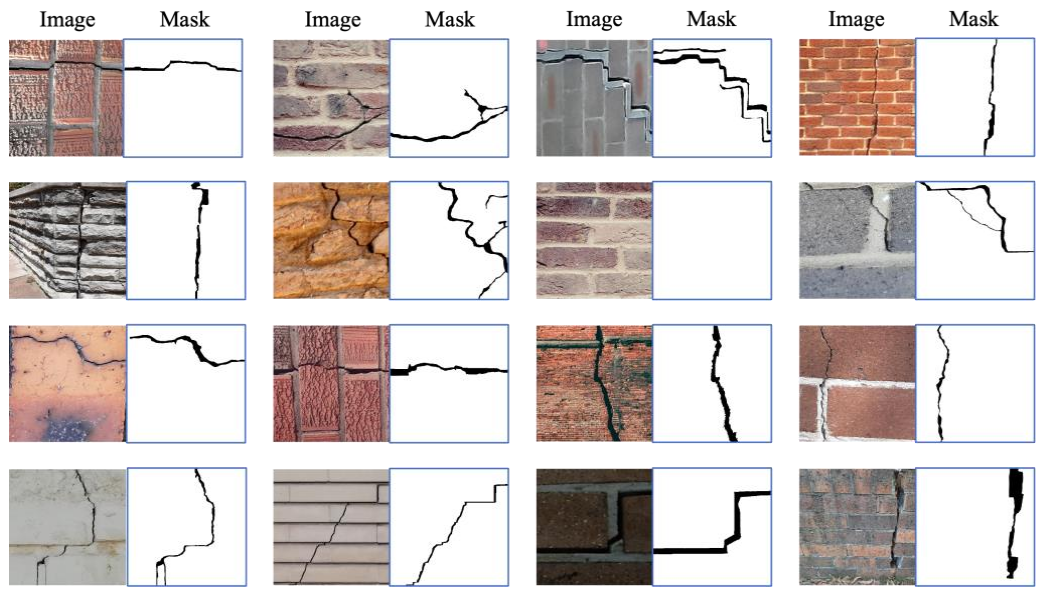

**Fig. 3** Image samples with their masks in dataset MC448 [18].

## 3.4 Training configuration

The proposed framework was implemented in Python with the PyTorch [53] open-source platform. All the validation and evaluation experiments were performed on a workstation featuring an AMD Ryzen 9 5900X 12-Core CPU, an NVIDIA GeForce RTX 3090 Ti GPU, and 32 gigabytes of RAM. The learning rate was configured to 0.001, utilising a batch size of two images, and the training progress for knowledge distillation lasted for 500 epochs employing the AdamW optimiser. To improve the generalisation of the training models and avoid over-fitting, colour-jittering and horizontal flipping were applied for training image augmentation.

## 3.5 Study framework

As illustrated in Fig. 4, the proposed RFKD framework was first compared with the SOTA KD methods to verify its novel attributes of knowledge transferring accuracy and robustness. The efficiency of RFKD with another CNN-based student model is demonstrated in Appendix-A.1. Then, the proposed lightweight model PCT with 0.5 M parameters was further compared with the existing representative lightweight models which have been deployed on edge devices, to evaluate the possibility of its practical deployment. In the first stage, three representative KD strategies (NKD, CWD, DIST) were chosen for the knowledge transferring accuracy and robustness comparison. In the second stage of the proposed PCT model assessment, all the representative model backbones were pretrained in the ImageNet [54] dataset to improve their final performances. In addition, mIoU and mDS metrics were utilised to evaluate the accuracy of crack segmentation, whereas model parameters, floating point operations (FLOPs) and inference time were considered to appraise the model efficiency. Four common image noises including pepper, salt, crack texture and gaussian blur were also used to investigate the model robustness against noise.

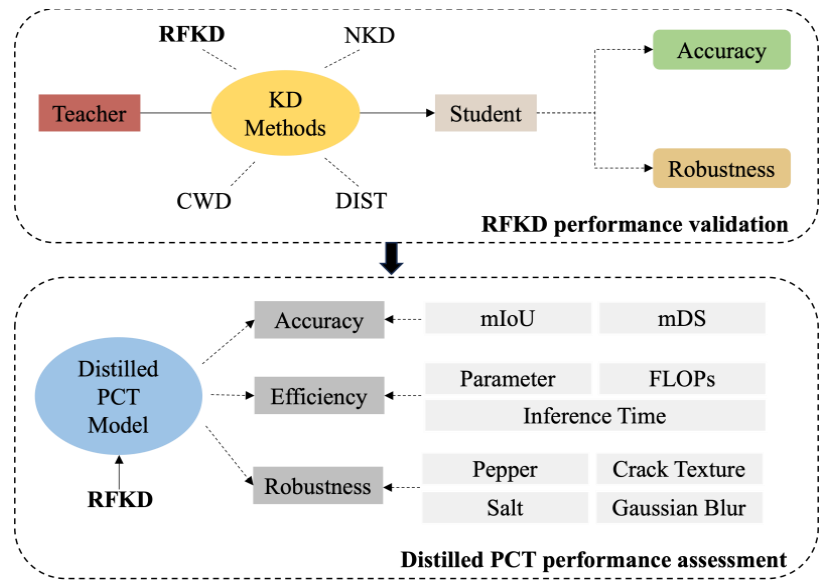

**Fig. 4** The schematic of the study framework.

## 4. Results and discussion

In this section, the proposed RFKD framework was first compared to the representative knowledge distillation methods and validated for its hypothetical performance. Then, the proposed PCT lightweight model trained with the help of RFKD, is compared to the commonly used lightweight model on edge devices and verified for its practical deployment.

**4.1 RFKD framework validation**

To fairly compare the performance of different knowledge distillation frameworks, the teacher model and student model were kept identical for each framework implementation, with the same training configuration. In this study, PoolingCrack with 32 M parameters and PCT with 0.5 M parameters were chosen as the teacher and student models, respectively.

4.1.1 Accuracy comparison

As illustrated in Table 1, the distilled student models with different KD methods are compared with the student model trained from scratch in terms of the model segmentation accuracy. The trained student model with the proposed RFKD method presents the best precision, which has 0.2% - 2.2% higher mDS and 1.1% - 2.9% higher mIoU than the other models. In addition, compared to training the student model from scratch, training the student model with different KD methods exhibits a significant improvement of 8.8% - 10.0% higher mDS and 8.6% - 11.5% mIoU. Since the students acquire knowledge from the teacher model, they also learn the improved feature representations and could produce more accurate segmentation results.

**Table 1** Quantitative accuracy comparison for the different KD methods.

| Knowledge distillation methods | Student model accuracy | |
|---|---|---|
| | mDS (%) | mIoU (%) |
| Student from scratch | 58.0 | 46.2 |
| NKD | 67.1 | 55.5 |
| CWD | 66.8 | 54.8 |
| DIST | 68.8 | 56.6 |
| **RFKD** | **69.0** | **57.7** |

Fig. 5 shows the ground truth and segmentation results of the masks produced by the distilled student models trained with different KD methods. Overall, the distilled student with the RFKD method presents the best ability to segment cracks over the complex masonry background texture. Compared with the student model trained from scratch, KD methods improve the crack pixel identification accuracy with more correctly classified crack pixels. Yet, the distilled students with NKD, CWD and DIST methods produce some visual noises as they generate the crack maps, in contrast to the distilled student with RFKD methods. The main reason for the improved segmentation performance of the distilled student under the RFKD method would be attributable to the additional feature learning (extra loss functions for controlling the learning process) of intermediate layers from the teacher model [55]. Since more accurate multiple-scale features are learned from the teacher model, the student model could generate more exact predictions in the final feature fusion stage, resulting in enhanced crack segmentation quality.

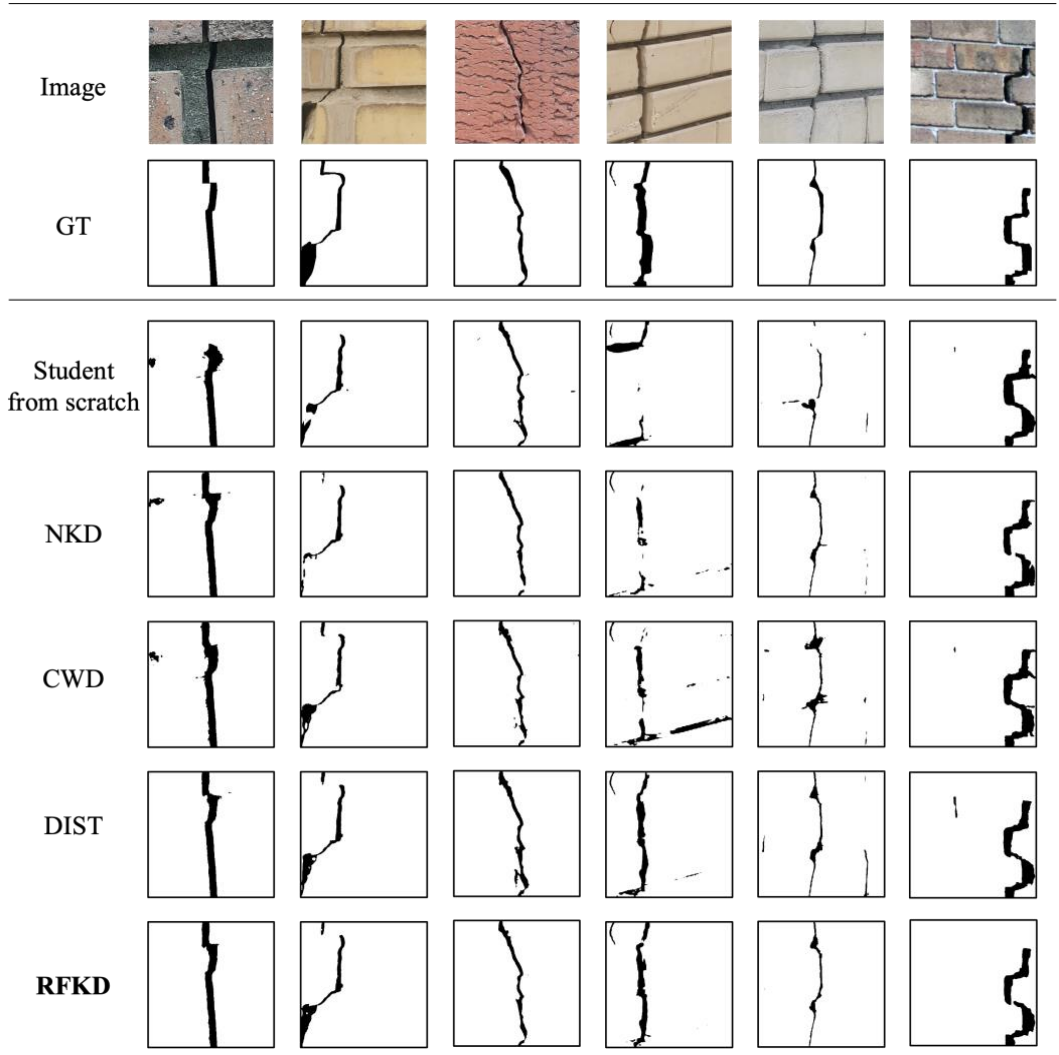

**Fig. 5** Qualitative accuracy comparison for the different KD methods.

4.1.2 Robustness comparison

To achieve automatic crack detection in a practical environment with different kinds of challenging backgrounds and input image signal interference, the model should be robust against various noise disturbances and exhibit stable performance [18]. Since the previous KD methods [31,32] for crack segmentation are not able to improve the robustness of the student model, the proposed RFKD method is aimed at addressing this issue. To validate the robustness of the RFKD method, four types of image noises [17,56], including crack texture, gaussian blur, and salt and pepper noises, were selected for the performance evaluation. Fig. 6 presents the robustness comparison of the distilled student models employing different KD strategies. The distilled student with the proposed RFKD method exhibits the best robustness, with up to a significant 62% higher mDS than the competing alternatives under these four noise disturbances. Notably, compared to the student model trained from scratch, while the distilled student models through the representative KD methods (NKD, CWD, DIST) could enhance mDS under low-intensity noise interference, their accuracy drops below that of the student from scratch as the noise intensity increases. However, the proposed RFKD method improves the mDS under all noise intensities, revealing its robust knowledge-transferring performance. Since the RFKD method utilises the mixed noisy and clean image inputs for training, it also gives the student the chance to learn robust features from the teacher, contributing to its enhanced anti-noise robustness.

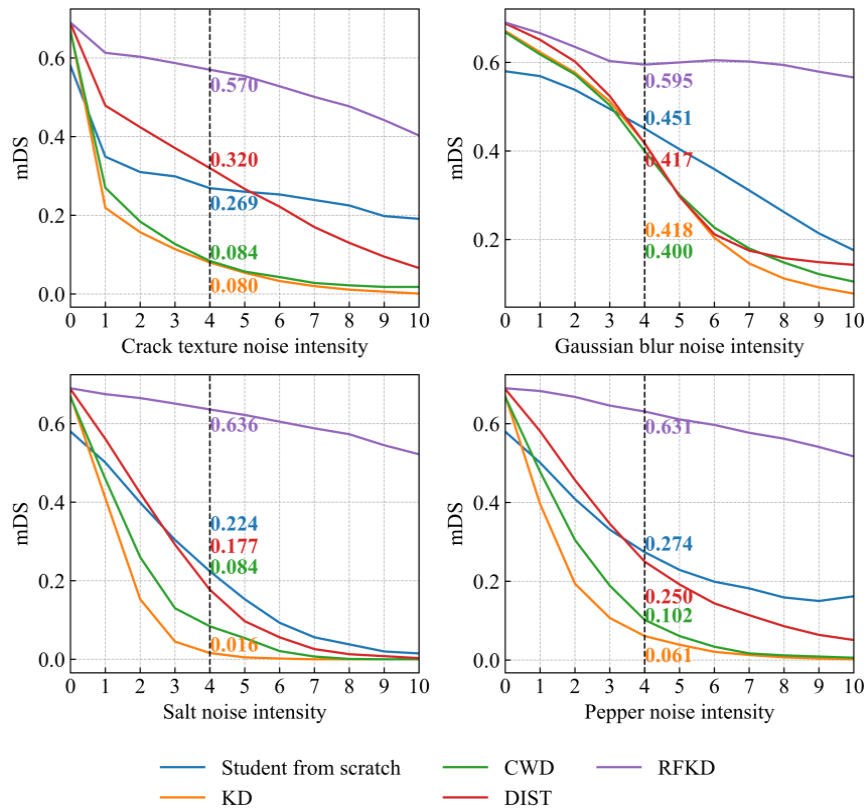

**Fig. 6** Quantitative robustness comparison for the different KD methods.

Fig. 7 contains the segmentation results generated by the distilled students with different KD methods on crack images distorted with 40% noise intensity. The distilled student with the RFKD method portrays the most complete crack profile in contrast to the others. Compared to the student trained from scratch, the RFKD distilled student exhibits significant improvement in accurate crack mask generation, while the distilled students with other KD methods have unstable performances. Overall, it appears that the proposed RFKD could effectively enhance the student model's accuracy and robustness.

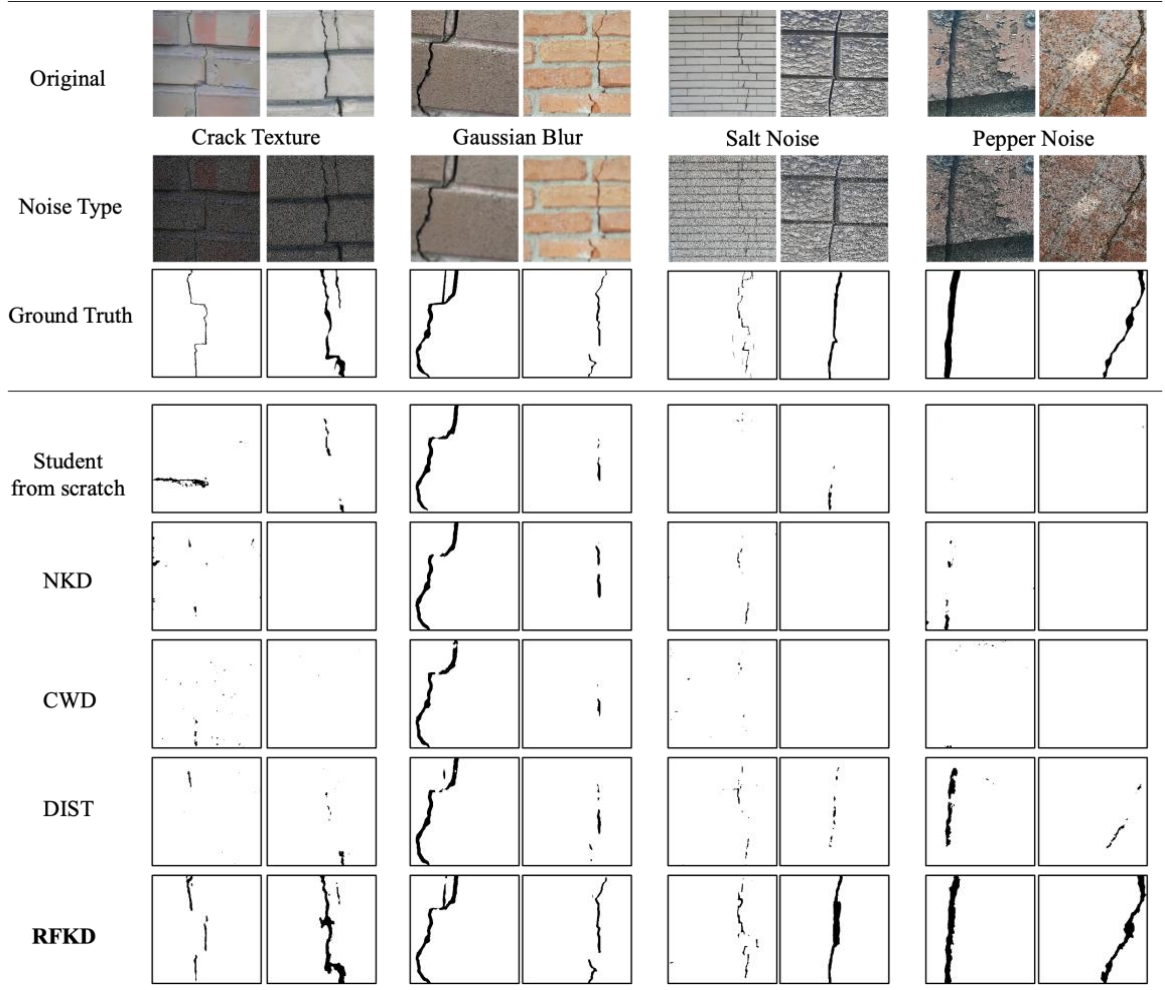

**Fig. 7** Qualitative robustness comparison for the different KD methods.

## 4.2 Efficiency assessment of the PCT model

To evaluate the ability of the RFKD distilled PCT model for practical deployment, currently established and SOTA representative lightweight models trained with transfer learning were selected for comparison.

### 4.2.1 Accuracy and efficiency comparison

The results of the accuracy and efficiency assessment are presented in Table 2 and Fig. 8. Overall, MobileOne and the PCT model could show the best accuracy on clean data. However, although the distilled PCT model presents the same mDS and 0.1% lower mIoU than MobileOne, the PCT model has 74% - 92% fewer parameters, 4% - 75% fewer FLOPs, and 16% - 74% less inference time than all others including MobileOne. It is also worth mentioning that the robustness analysis presented in Fig. 9 further reveals the significantly improved performance of the PCT model on noisy data. Since there is a trade-off between the model scale and the prediction accuracy, the main aim is not only to design a model with high accuracy on clean data but also to develop a light model with high generalisation robustness against noisy data for practical deployment, as in real-world case scenarios many unprecedented conditions may occur.

Table 2 Quantitative accuracy comparison for the different lightweight models

| Model | Parameters (M) | FLOPs (G) | mDS (%) | mIoU (%) | Inference time (ms) |
|---|---|---|---|---|---|
| MobileOne | 6.3 | 10.9 | **69.0** | **57.8** | 21.5 |
| MobileNetV2 | 4.2 | 7.5 | 68.5 | 57.2 | 6.7 |
| EfficientNet-B0 | 5.8 | 7.8 | 68.9 | 57.5 | 11.1 |
| MobileViT | 1.9 | 2.8 | 62.0 | 49.7 | 7.9 |
| **PCT** | **0.5** | **2.7** | **69.0** | 57.7 | **5.6** |

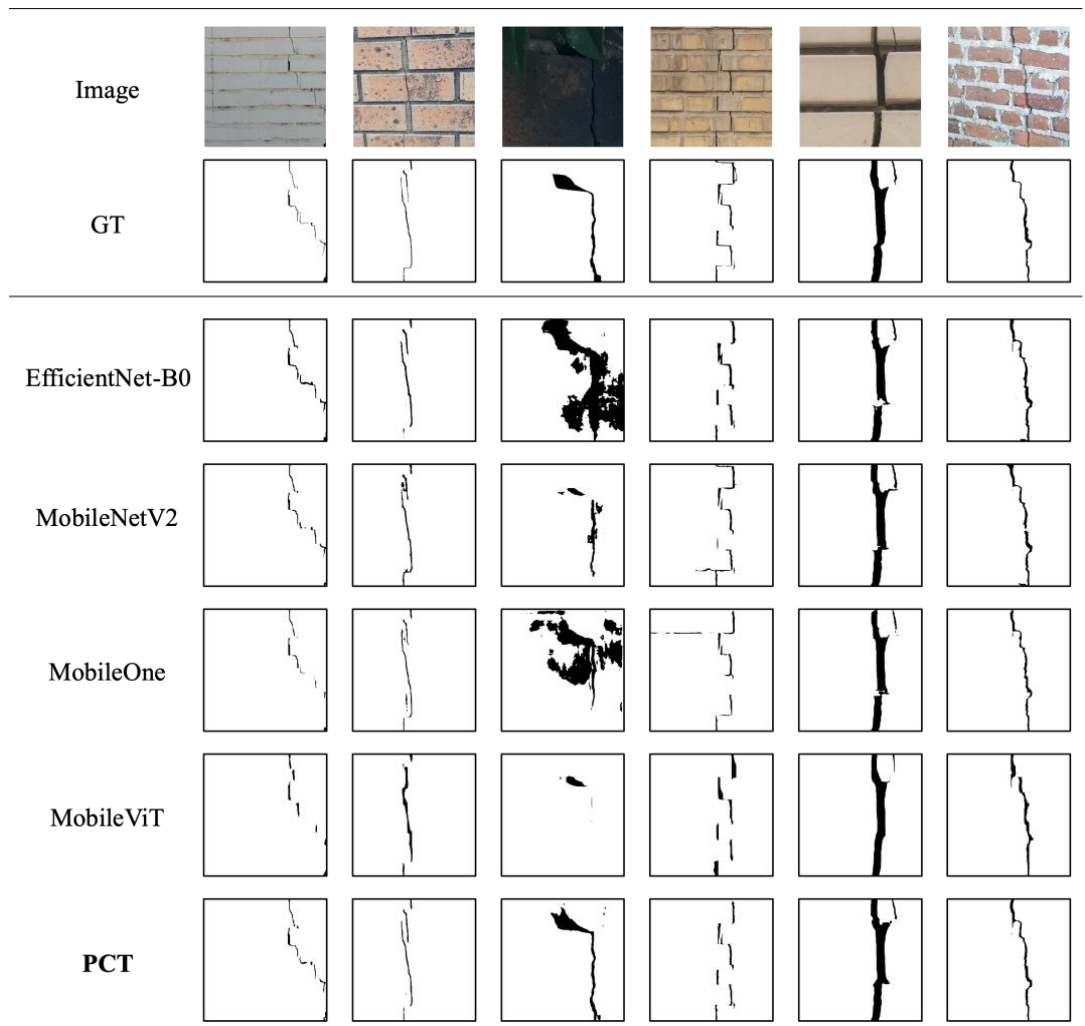

Fig. 8 Qualitative accuracy comparison for the different lightweight models.

As shown in Fig. 8, the segmentation results of different lightweight models in terms of the randomly selected crack images are visualised. Overall, the segmented crack profiles by the distilled PCT demonstrate comparable completeness in contrast to other lightweight models. Yet, when one crack image in the third column is under low luminance conditions, the distilled

PCT maintains stable segmentation performance, while other lightweight models cannot provide accurate segmentation results. This difference would be attributed to the inferior robustness of those other lightweight models and the robust features learned by the distilled PCT.

4.2.2 Robustness comparison

As presented in Fig. 9, the distilled PCT model presents the best crack segmentation precision under the interferences of all types of image noises, with up to 59.2% higher mDS than the other ones. While the other lightweight models could have considerable accuracy and provide consistent crack profile predictions without noise disturbances, their performances drop as the noise intensity increases. Fig. 10 shows the segmented results of 40% noise intensity crack images for different lightweight models. The distilled PCT generates more precise and consistent crack masks than the competitive models. Due to the inconsistent performance under noise interferences, the other models fail to portray the complete crack profile or sometimes produce several false crack pixels. The distilled PCT model learns more robust crack feature representations under the supervision of the teacher model through the proposed RFKD method, contributing to its stable performance under noise interruptions.

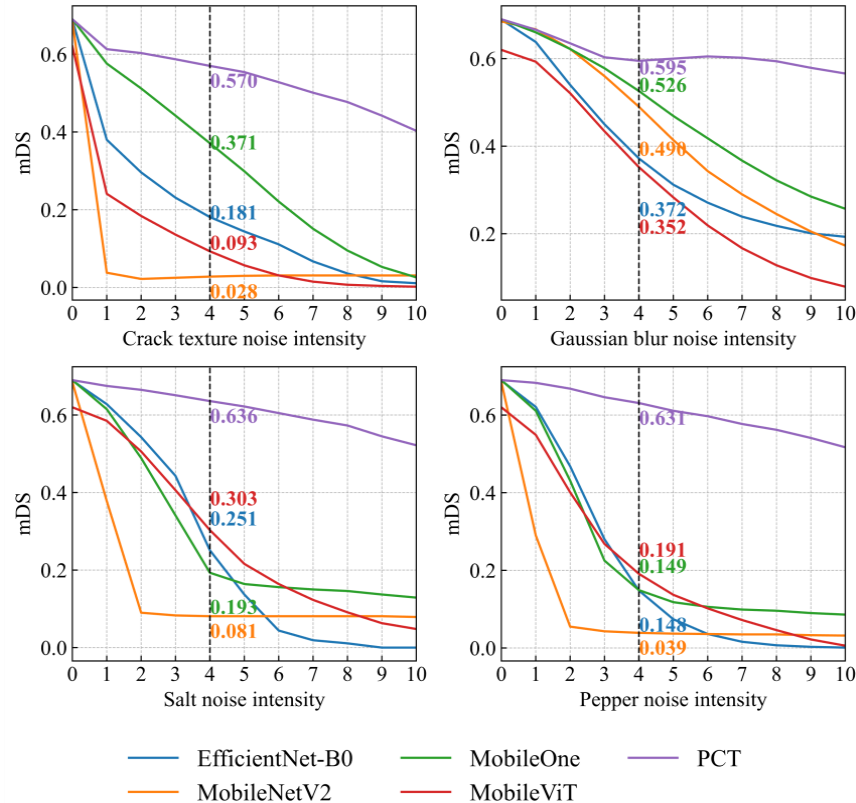

**Fig. 9** Quantitative robustness comparison for the different lightweight models.

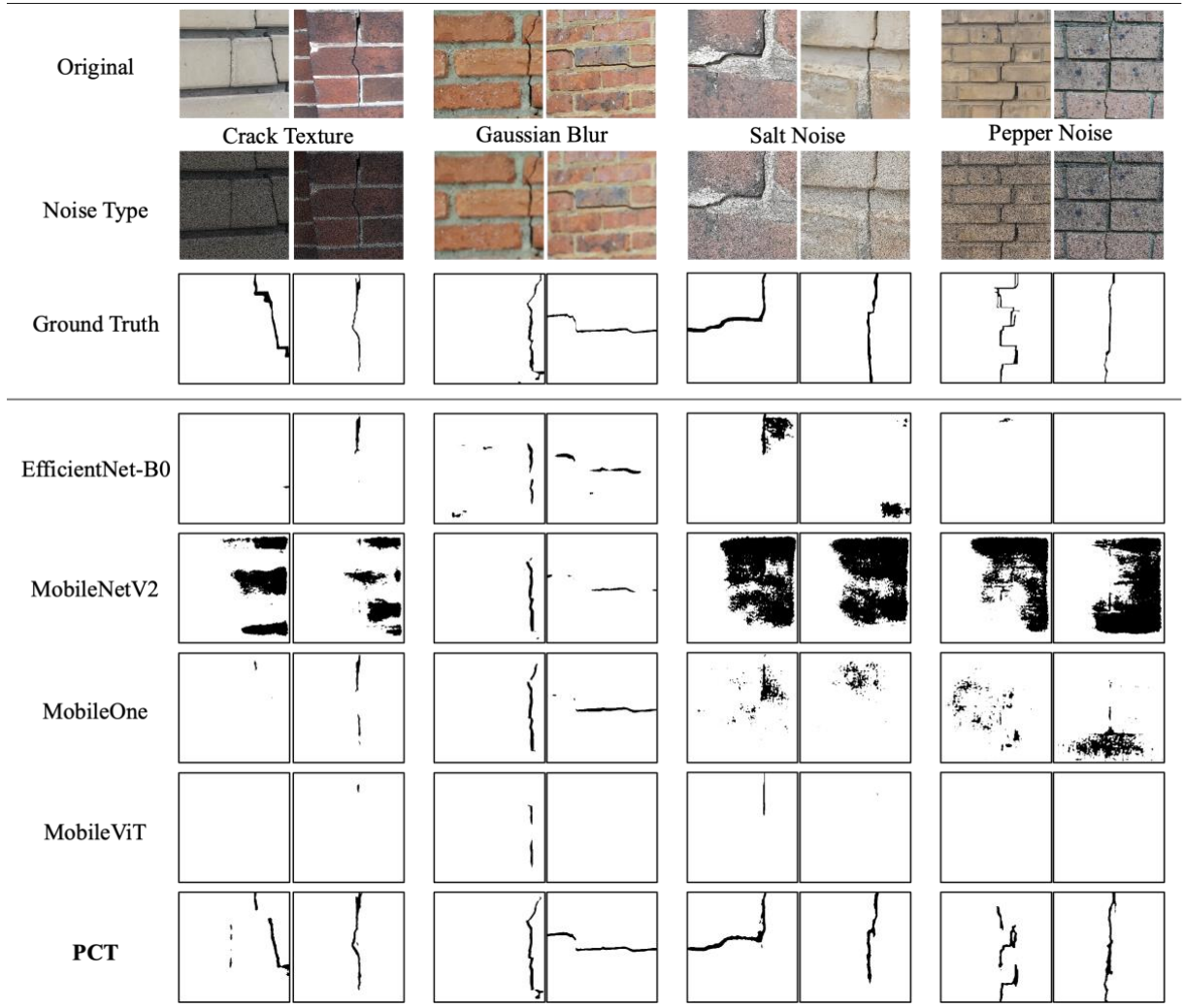

Fig. 10 Qualitative robustness comparison for the different lightweight models.

## 5. Conclusion

Since current knowledge distillation methods for crack segmentation cannot improve both the accuracy and robustness of the student models, a novel framework named RFKD is proposed to address this issue. The method enforces the student model to learn features of the teacher model in both intermediate and logit layers through the input of mixed noisy and clean training images. The primary assumption is that this framework can assist the student in learning the robust features from the teacher and contribute to both the precision and robustness improvement of the student.

To verify the performance of the proposed RFKD method, a validation comparison was conducted on the MC448 dataset with the masonry background. The results showed that RFKD could assist the student model in obtaining 0.2% - 2.2% higher mDS and 1.1% - 2.9% higher mIoU than the previous NKD, CWD, and DIST methods on the clean data, and up to 62% higher mDS on the noisy data. In addition, to demonstrate the further application of the RFKD, a lightweight Transformer-based student model PCT was designed and compared with the current representative lightweight models. The comparative outcomes indicated that the PCT model distilled by RFKD could achieve identical mDS and slightly lower mIoU (0.1%) than the most accurate and largest MobileOne model on the clean data, while up to 59.2% higher mDS than other representative lightweight models on the noisy data, with 74% - 92% fewer parameters, 4% - 75% fewer FLOPs, and 16% - 74% less inference time.

As the developed RFKD framework was verified to provide enhanced precision and robustness performance for the lightweight model, it paves the way for future lightweight model designs and relevant practical deployments on edge devices.

**Declaration of Competing Interest**

The authors declare that they have no conflict of interests for the work in this paper.

**Acknowledgement**

The authors acknowledge the Sydney Informatics Hub and the use of the University of Sydney's high-performance computing cluster, Artemis. The authors acknowledge the support from the University of Sydney through the Digital Sciences Initiative program.

**Appendix A**

**A.1 RFKD for another CNN-based lightweight student model**

Since the proposed RFKD has been demonstrated available for the designed Transformer-based PCT lightweight student model, another validation is conducted to assess its general performance for other student models. The CNN-based UNet-Resnet18 model [57] was chosen for this verification, utilising the same training configuration and RFKD strategy as the PCT model. Table 3 and Fig. 11 illustrate the accuracy and robustness comparison of the UNet-ResNet18 and PCT students.

Table 3 Accuracy comparison between UNet and PCT under the RFKD strategy.

| Student Model | Parameters (M) | Training from scratch | | RFKD | |
|---|---|---|---|---|---|
| | | mDS (%) | mIoU (%) | mDS (%) | mIoU (%) |
| UNet-ResNet18 | 14.3 | 61.8 | 50.7 | 68.1 (+6.3) | 57.2 (+6.5) |
| PCT | 0.5 | 58.0 | 46.2 | 69.0 (+11.0) | 57.7 (+11.5) |

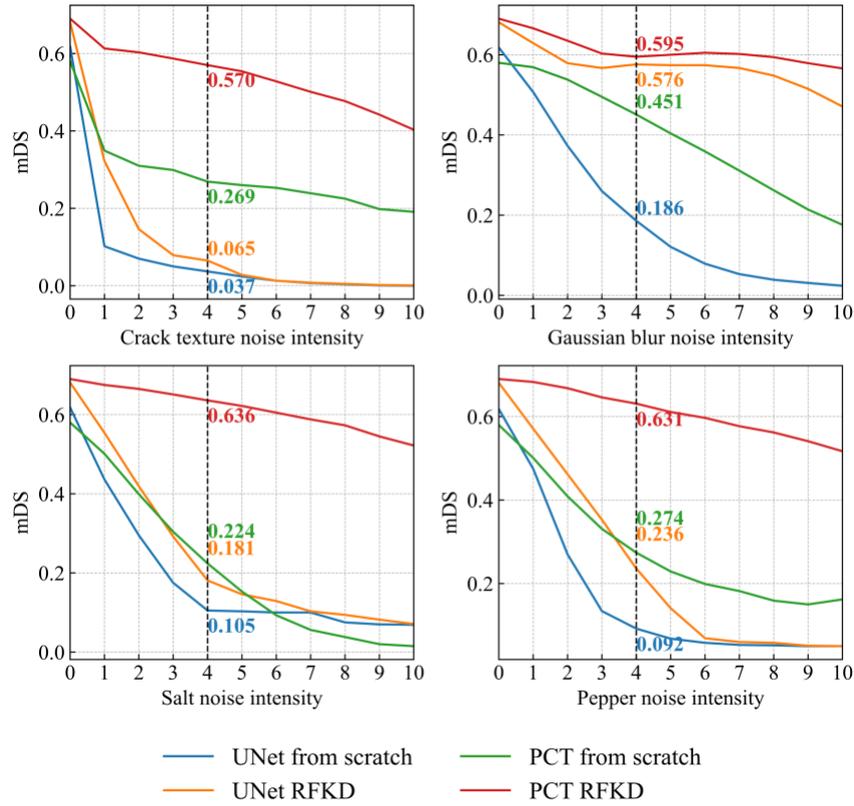

**Fig. 11** Robustness comparison between UNet and PCT under the RFKD strategy.

As presented in Table 3 and Fig. 11, the proposed RFKD method improves the accuracy and robustness of the UNet-ResNet18 model. Thus, the general utilisation of the RFKD method for different lightweight student models is demonstrated. Yet, in Fig. 11, the enhanced robustness of UNet-ResNet18 is less significant than that of the PCT model, especially for the robustness against crack texture, salt, and pepper noises. Due to the limitation of the fixed-size convolutional kernel and local attention [58], CNN-based models would be weaker than similar scale Transformer-based models when processing noisy images. This instinctive shortcoming of UNet-ResNet18 restricts the further improvement of its robustness through the RFKD method.

**A.2 The architecture of the teacher model**

Since the PoolingCrack model [42] has demonstrated advanced performance in several crack image datasets, it was chosen as the teacher model in this study. Fig. 12 and Fig. 13 present the overall and detailed structure of the PoolingCrack model. This model consists of four components, including convolutional patch embedding [59], pooling blocks [60], feature selection modules [44] and feature alignment modules [45], with four down-sampling and up-sampling stages. The details of the pooling blocks, feature selection modules and feature alignment modules are introduced in Fig. 12. Further details about the model can be found in [42].

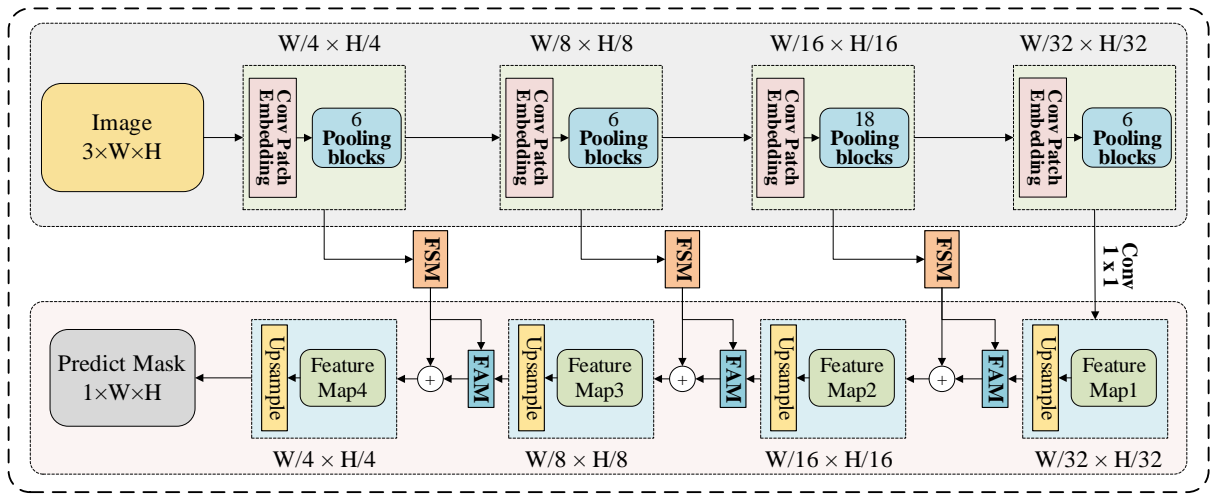

**Fig. 12** The overall architecture of the PoolingCrack model.

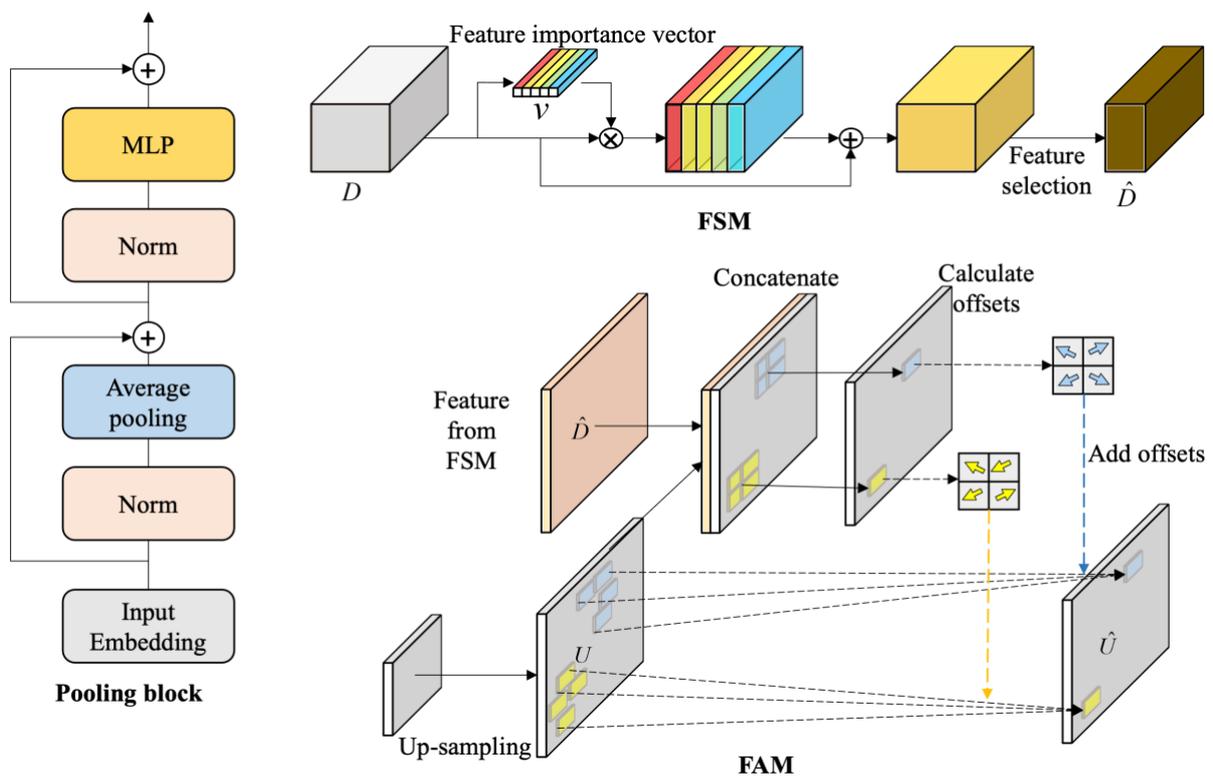

**Fig. 13** The details of the component parts in the PoolingCrack model.